\documentclass{article}

\usepackage[preprint,nonatbib]{neurips_2024}

\usepackage[utf8]{inputenc} % allow utf-8 input
\usepackage[T1]{fontenc}    % use 8-bit T1 fonts
\usepackage{hyperref}       % hyperlinks
\usepackage{url}            % simple URL typesetting
\usepackage{booktabs}       % professional-quality tables
\usepackage{amsfonts}       % blackboard math symbols
\usepackage{nicefrac}       % compact symbols for 1/2, etc.
\usepackage{microtype}      % microtypography
\usepackage{xcolor}         % colors
\usepackage{algorithm}
\usepackage{graphicx}
\usepackage{algorithmic}

\title{Small or Large? Zero-Shot or Finetuned? Guiding Language Model Choice for Specialized Applications in Healthcare}

\author{%
  Lovedeep Gondara\thanks{lovedeep.gondara@phsa.ca} \\
  British Columbia Cancer Registry\\
  Provincial Health Services Authority\\
  Vancouver, Canada \\
   \And
    Jonathan Simkin \\
  British Columbia Cancer Registry\\
  Provincial Health Services Authority\\
  Vancouver, Canada \\
  \And
    Graham Sayle \\
  Data Science Institute\\
  University of British Columbia\\
  Vancouver, Canada \\
    \And
    Shebnum Devji \\
  British Columbia Cancer Registry\\
  Provincial Health Services Authority\\
  Vancouver, Canada \\
  \And
    Gregory Arbour \\
  Data Science Institute\\
  University of British Columbia\\
  Vancouver, Canada \\
   \And
    Raymond Ng \\
  Data Science Institute\\
  University of British Columbia\\
  Vancouver, Canada \\
}

\begin{document}

\maketitle

\begin{abstract}
This study aims to guide language model selection by investigating: 1) the necessity of finetuning versus zero-shot usage, 2) the benefits of domain-adjacent versus generic pretrained models, 3) the value of further domain-specific pretraining, and 4) the continued relevance of BERT-style, bidirectional Language Models (BiLMs) compared to modern Large Language Models (LLMs) for specific tasks. Using electronic pathology reports from the British Columbia Cancer Registry (BCCR), three classification scenarios with varying difficulty and data size are evaluated. Models include various BiLMs and an LLM. BiLMs are evaluated both zero-shot and finetuned; the LLM is evaluated zero-shot only. Finetuning significantly improved BiLMs performance across all scenarios compared to their zero-shot results. The zero-shot LLM outperformed zero-shot BiLMs but was consistently outperformed by finetuned BiLMs. Domain-adjacent BiLMs generally performed better than the generic BiLM after finetuning, especially on harder tasks. Further domain-specific pretraining yielded modest gains on easier tasks but significant improvements on the complex, data-scarce task. The results highlight the critical role of finetuning for BiLMs in specialized domains, enabling them to surpass zero-shot LLM performance on targeted classification tasks. Pretraining on domain-adjacent or domain-specific data provides further advantages, particularly for complex problems or limited finetuning data. While LLMs offer strong zero-shot capabilities, their performance on these specific tasks did not match that of appropriately finetuned BiLMs. In the era of LLMs, BiLMs remain relevant and effective, offering a potentially superior performance-resource trade-off compared to LLMs.
\end{abstract}

\section{Introduction}
The rapid advancement and proliferation of language models (LMs) have revolutionized the field of Natural Language Processing (NLP) \cite{qin2024large}. From compact, BERT-style, bidirectional language models (BiLMs) \cite{devlin2018bert} to modern, massive large language models (LLMs) like GPT-4, practitioners now have more choices than ever. This raises a critical question: in the era of language models, which LM should be used for a given task? The decision involves several considerations.

\begin{enumerate}
    \item Do we need to finetune a model or can we use it in its zero-shot capacity?
    \item Are domain-adjacent pretrained models better than generic pretrained models?
    \item Is further pretraining on domain specific data helpful?
    \item With the rise of LLMs, are BiLMs like BERT still relevant?
\end{enumerate}

This paper delves into these crucial decision points and explores the various trade-offs. For empirical evidence, we use the electronic pathology data from the British Columbia Cancer Registry (BCCR) and study various scenarios that help us answer the questions above. 

Scenario a) Easy problem, binary classification, large training data. For this scenario, we use the reportability classification problem \cite{gondara2024classifying}, where the training data consists of 40,000 labelled pathology reports with the goal being to classify the reports into reportable or non-reportable tumors (based on the guidelines set by governing bodies). The test data for this scenario consists of 20,400 pathology reports unseen by the models during the training phase. 

Scenario b) Medium hard problem, multi-class classification, limited training data. For this scenario, we use the tumor group classification problem \cite{gondara2025elm}, where the training data consists of 16,000 pathology reports with the goal being to classify the pathology reports into one of nineteen tumor groups, where most tumor groups have less than 1000 samples in the training data. The test data for this scenario consists of 2,058 pathology reports.

Scenario c) Hard problem, multi-class classification, small training data. This scenario is based on classifying histology from pathology reports for leukemia, where we have the six most common histology codes and only 1,000 pathology reports as the training data and 447 reports are used as the test data.

These scenarios were selected to represent a spectrum of common challenges in clinical NLP, varying in task complexity (binary vs. multi-class) and data availability (large vs. limited vs. small)

\section{Methods and Metrics}
For empirical evaluation, we use several off-the-shelf BiLMs: a strong general-purpose model like RoBERTa (125M parameters) \cite{liu2019roberta}, and models specifically pre-trained on broader clinical text (pathologyBERT (108M parameters) \cite{santos2022pathologybert}, and Gatortron (345M parameters) \cite{yang2022gatortron}). For our LLM, we use Mistral nemo instruct (12B parameters)\footnote{\url{https://huggingface.co/nvidia/Mistral-NeMo-12B-Instruct}} \cite{jiang2023mistral} selected as a representative high-performance open-source LLM. We further pretrain RoBERTa and Gatortron on 1M pathology reports from BCCR using Masked Language Modeling (MLM) and denote these domain specific pretrained models as BCCRoBERTa and BCCRTron. For performance reporting, we use the macro-averaged F1 scores.

\section{Off-the-shelf Models: Finetuning vs. Zero-Shot }
A fundamental decision when deploying a pre-trained language model concerns the adaptation strategy. Should we invest resources in finetuning the model on a task-specific dataset, or can the model be used effectively "out-of-the-box" using zero-shot approaches? Large Language Models (LLMs), particularly those with billions of parameters, have demonstrated remarkable zero-shot and few-shot capabilities \cite{brown2020language}. By providing task instructions and, optionally, a few examples directly within the input prompt, these models can often perform surprisingly well on tasks for which they weren't explicitly trained. This approach is advantageous as it requires minimal or no task-specific labeled data and can significantly speed up development cycles. However, zero-shot performance can be variable and highly sensitive to prompt phrasing, and may not reach the peak accuracy achievable through dedicated training for complex or highly specialized tasks. Finetuning, alternatively, involves updating the model's weights using a labeled dataset specific to the target task \cite{howard2018universal}. This typically leads to higher performance on the specific task, better adaptation to domain-specific nuances, and potentially more reliable outputs. The trade-offs include the need for labeled data and the computational cost of the training process.

For empirical evaluation using our scenarios, we observe that for BiLMs, finetuning consistently significantly outperforms the zero-shot approaches where zero-shot performance ranges from 0.34-0.40 for scenario (a), 0.01 for scenario (b), and 0.02-0.13 for scenario (c). After finetuning, the performance increased to 0.95-0.97 for scenario (a), 0.78-0.85 for scenario (b), and 0.60-0.78 for scenario (c). For LLM, in its zero-shot capacity, it reached the performance of 0.76 for scenario (a), 0.54 for scenario (b), and 0.65 for scenario (c). LLM zero-shot performance was evaluated using task-specific instructions provided via prompting. To mirror resource constraints encountered in the real-world applications, we do not finetune the LLM. We provide detailed results in a Table \ref{tab:detailed_results}. The results align with expectations, as BiLMs lack the zero-shot capabilities of modern LLMs, particularly for out-of-domain tasks.

\section{The Value of Further Domain-Specific Pretraining}
The question of whether further pretraining on domain-specific data is beneficial often arises, especially when base models are already trained on broad or domain-adjacent datasets. For instance, if a model like Gatortron or pathologyBERT has been pre-trained on a large corpus of clinical notes, is there value in continuing the pretraining process specifically on the data from one's own institution before finetuning for a downstream task? The argument for further pretraining rests on the potential for the model to learn domain-specific vocabulary, jargon, syntax, and subtle semantic relationships that might be underrepresented or absent in the initial pretraining corpus \cite{gururangan2020don}. Further pretraining on specific data allows the model to adjust its internal representations to better reflect this unique linguistic distribution, potentially leading to improved performance on downstream tasks like information extraction or classification within that narrow domain.

However, this process requires access to a substantial amount of unlabeled domain-specific text and incurs significant computational costs. The benefits may also diminish if the initial pretraining data was already highly relevant or if the downstream task dataset is large enough for finetuning to effectively adapt the model. Therefore, the decision to pursue further pretraining requires a careful cost-benefit analysis, weighing the potential performance gains against the required resources and the degree of divergence between the existing model's knowledge and the target domain's specific characteristics.

Using BCCRoBERTa and BCCRTron, for scenarios (a) and (b), we observe that finetuned BCCRoBERTa outperforms finetuned RoBERTa (0.97 vs 0.96 and 0.84 vs 0.78 respectively), whereas BCCRTron shows no improvement compared to Gatortron.  This suggests that for these tasks (a and b) with relatively larger fine-tuning datasets, Gatortron's initial large-scale clinical pretraining may have already captured sufficient domain-relevant features, while the general-purpose RoBERTa benefited more from pathology-specific adaptation. For scenario (c), we observe significant differences when using further pretrained models (0.71 vs 0.61 for BCCRoBERTa vs RoBERTa and 0.89 vs 0.78 for BCCRTron vs Gatortron).

% \usepackage{booktabs}
\begin{table}[h]
\centering
\begin{tabular}{@{}llll@{}}
\toprule
Models                    & Scenario a & Scenario b & Scenario c \\ \midrule
RoBERTa - zeroshot        & 0.34                                  & 0.01                                & 0.02                              \\
PathologyBERT - zeroshot  & 0.40                                  & 0.01                                & 0.04                              \\
Gatortron - zeroshot      & 0.34                                  & 0.01                                & 0.13                              \\
Mistral - zeroshot        & 0.76                                  & 0.54                                & 0.65                              \\
RoBERTa  - finetuned      & 0.96                                  & 0.78                                & 0.61                              \\
PathologyBERT - finetuned & 0.95                                  & 0.81                                & 0.60                              \\
Gatortron - finetuned     & 0.97                                  & 0.85                                & 0.78                              \\
BCCRoBERTa -finetuned     & 0.97                                  & 0.84                                & 0.71                              \\
BCCRTron -finetuned       & 0.97                                  & 0.85                                & 0.89                              \\ \bottomrule
\end{tabular}
\caption{Detailed results for all models and all scenarios}\label{tab:detailed_results}
\end{table}

\section{Revisiting the Questions}
With the help of the information provided above, we conclude the paper by revisiting the questions and providing our recommendations based on our experience and empirical evidence, while acknowledging the limitation that we have only tested a small set of models.

\begin{enumerate}
    \item \emph{Do we need to finetune a model or can we use it in its zero-shot capacity?} Our evaluation clearly shows that finetuned models significantly outperform their zero-shot counterparts, especially for BiLMs, as they have limited zero-shot capabilities. LLMs, as anticipated perform better in zero-shot capacity than their BiLM counterparts, but are significantly outperformed by finetuned BiLMs.
    
    \emph{Our recommendation:} BiLMs should be finetuned on the task-specific dataset. A finetuned BiLM outperforms a zero-shot LLM for many well-defined tasks, especially where the models are intended for use in a specialized domain (cancer pathology reports in our case).

    \item \emph{Are domain-adjacent pretrained models better than general pretrained models?}
    We compare the finetuned version of general pretrained models (RoBERTa) with domain-adjacent pretrained models (PathologyBERT and Gatortron), and we observe that the domain adjacent models, when finetuned often outperform general pretrained models, especially for complex tasks where the data availability for finetuning is scarce.
    
    \emph{Our recommendation:} Echoing the recommendations from \cite{raffel2020exploring}, where possible, finetune a domain-adjacent pretrained model compared to a general model.

    \item \emph{Is further pretraining on domain specific data helpful?}
    While further pretraining on domain specific data offers small gains for simple tasks, it significantly improves the downstream model performance for complex tasks and/or when the finetuning dataset is small (such as histology classification in our case).
    
    \emph{Our recommendation:} If the data and compute resources are available, it is worthwhile further pretraining BiLMs for extra performance boost.

    \item \emph{With the rise of LLMs, are BiLMs like BERT still relevant?}
    We have shown that for specialized tasks, BiLMs still endure and outperform zero-shot LLMs. While LLMs excel at generation, complex reasoning, and few-shot learning across a vast range of topics, BiLMs often provide sufficient or even superior performance for many well-defined NLP tasks, such as text classification, named entity recognition,etc. particularly when task-specific training data is available for finetuning. Further, for resource-constrained environments, which are a common place, BiLMs offer a compelling balance of performance and efficiency.
    
    \emph{Our recommendation:} If the task is well defined (example: classification) and the data is available for finetuning, BiLMs should be the first choice.
\end{enumerate}

\bibliographystyle{unsrt}
\bibliography{sample}

\end{document}